\documentclass[
]{ceurart}

\begin{document}

\copyrightyear{2022}
\copyrightclause{Copyright for this paper by its authors.
  Use permitted under Creative Commons License Attribution 4.0
  International (CC BY 4.0).}

\conference{In: R. Campos, A. Jorge, A. Jatowt, S. Bhatia, M. Litvak (eds.): Proceedings of the Text2Story'22 Workshop, Stavanger (Norway), 10-April-2022
}

\title{Simplifying Multilingual News Clustering Through Projection From a Shared Space}

\author{João Santos}[
email=joao.santos@priberam.pt,
]
\author{Afonso Mendes}[
email=amm@priberam.pt,
]
\author{Sebastião Miranda}[
email=sebastiao.miranda@priberam.pt,
]
\address{Priberam Labs, Alameda D. Afonso Henriques, 41, 2º, 1000-123 Lisboa, Portugal}

\begin{abstract}
The task of organizing and clustering multilingual news articles for media monitoring is essential to follow news stories in real time. Most approaches to this task focus on high-resource languages (mostly English), with low-resource languages being disregarded. With that in mind, we present a much simpler online system that is able to cluster an incoming stream of documents without depending on language-specific features. We empirically demonstrate that the use of multilingual contextual embeddings as the document representation significantly improves clustering quality. We challenge previous crosslingual approaches by removing the precondition of building monolingual clusters. We model the clustering process as a set of linear classifiers to aggregate similar documents, and correct closely-related multilingual clusters through merging in an online fashion. Our system achieves state-of-the-art results on a multilingual news stream clustering dataset, and we introduce a new evaluation for zero-shot news clustering in multiple languages. We make our code available as open-source.
\end{abstract}


\maketitle

\section{Introduction}

The last few decades have been characterized by an exponential growth in the amount of news sources available, with the task of following news stories in real-time becoming very difficult to perform manually. As such, a demand has risen for systems that are capable of monitoring and organizing articles into news stories. Most approaches to this task focus mainly on the English language \cite{newslens,dense_vs_sparse,entity_aware}, with multilingual systems being highly dependent on language-specific features such as the entities of a given document \cite{batch_clustering,miranda}. 
Those approaches perform poorly on a multilingual scenario and are hard to extend to low-resource languages. Taking those limitations into account, we propose an online news clustering system that is able to cluster documents across languages (for which there are pretrained multilingual contextual embeddings) while maintaining performance regarding monolingual scenarios.

The contributions described in the paper are: \textbf{(i)} We develop a system that is able to cluster documents without depending on language-specific features; \textbf{(ii)}  We empirically demonstrate that the use of multilingual contextual embeddings as the document representation significantly improves clustering quality;
    \textbf{(iii)} We propose a method to train a classifier in order to merge similar clusters in an online setting, and demonstrate its impact in obtaining state-of-the-art results for multilingual clustering;
    \textbf{(iv)} We show that our system performs well on languages not seen during training and we describe a zero-shot experimental setting for Chinese, Russian, French, Italian, Slovenian and Croatian.

\section{Related Work}
The Topic Detection and Tracking (TDT) task \cite{tdt} has the goal of, given a stream of news articles, to arrange the documents into topic clusters called stories.

Regarding batch clustering approaches, Laban et al. introduce \textit{newsLens} \cite{newslens}, a batch-based approach to news clustering. \textit{NewsLens} constructs its stories by extracting keywords from the articles and linking them through a community detection algorithm. Staykovski et al. \cite{dense_vs_sparse} follow \textit{newsLens}'s work by implementing a sparse approach through TF-IDF bag-of-words document representations, and compare it against a doc2vec \cite{doc2vec} dense representation approach. Linger et al. \cite{batch_clustering} extend the aforementioned studies into a crosslingual setting by processing batches of articles into monolingual topics and using a fine-tuned multilingual DistilBERT \cite{distilbert} to link topics across languages.

For online clustering, Miranda et al. \cite{miranda} approach the problem by processing a continuous stream of multilingual documents into monolingual and crosslingual clusters. Each document is first associated to a monolingual cluster through sparse features, and crosslingual clusters are computed by linking different monolingual clusters using crosslingual word embeddings \cite{gardner}. Saravanakumar et al. \cite{entity_aware} propose an online news clustering system based on the non-parametric K-means algorithm. Their approach uses both sparse and dense features, with a main emphasis on the use of a fine-tuned entity-aware BERT model \cite{bert_base} to produce dense document representations, and is evaluated for the English language.

Our system follows the described online clustering approaches to the TDT task: while Miranda et al. was bound to the usage of specific individual models for each language due to processing monolingual clusters, our approach provides a system that is able to leverage on dense multilingual document representations without the need to process the documents into monolingual clusters first. This is accomplished by using a single crosslingual representation for the documents and the consequent training of our ranking and classification models at a crosslingual level, which also allows for a fully dense clustering space and guarantees that our system is not limited to the English language, unlike Sarvanakumar et al.'s approach.
\begin{comment}
explanation for "much simpler system"
\end{comment}

\section{The Clustering Algorithm}

\begin{figure*}[h]
\centering
\includegraphics[width=\textwidth]{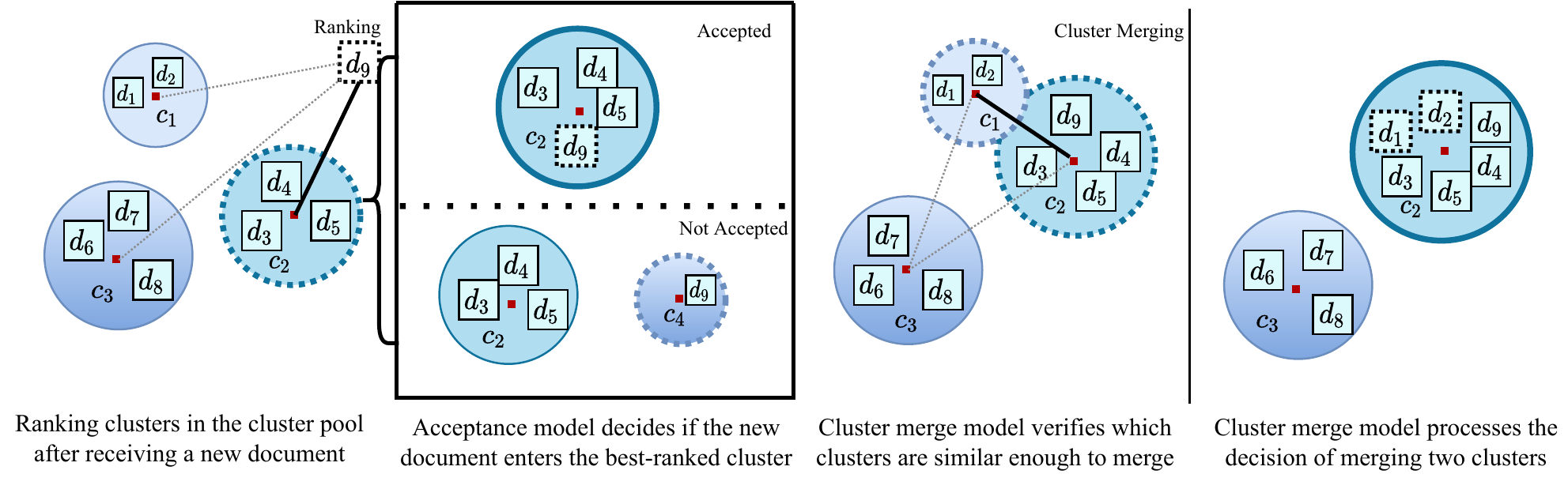}
\caption{Representation of our clustering system's  ranking, acceptance and merge steps.}
\label{fig:architecture}
\end{figure*}

Our main focus for this task is to build an online multilingual news clustering system that depends as little as possible on language-specific features in order to process news articles for zero-shot languages (where we have no clustering training data) without having a considerable loss in performance. Previous approaches mostly focus on a single language \cite{newslens,dense_vs_sparse,entity_aware} or a specific set of languages \cite{batch_clustering,miranda}.
\begin{comment}
In particular, extracting information about entities can be a challenging task for low-resource languages. 
\end{comment}

Our system is composed by four main steps (partially displayed in Figure \ref{fig:architecture}): obtaining the document representations, computing the best-ranked cluster, deciding if the document accepts the best-ranked cluster and enters it, and merging clusters that pertain to the same story.

\subsection{Document Representation}
\label{doc_repr}

In contrast to previous work, we use a representation for each document that does not depend on its language, thus eliminating the need of distinguishing representations between monolingual and crosslingual.  Documents are comprised of two components: a set of dense vectors $d^r$ corresponding to a contextual representation of the document, and a temporal representation $d^{ts}$. To obtain the dense vectors we use \textit{distiluse}\footnote{https://huggingface.co/sentence-transformers/distiluse-base-multilingual-cased-v2}, a model produced through knowledge distillation \cite{knowledge_distil} on Multilingual DistilBERT \cite{distilbert} by using mUSE (multilingual Universal Sentence Encoder) \cite{muse} as its teacher model. Similarly to mUSE, this model aligns text at the sentence-level \cite{sentencebert} into a shared semantic space, thus, similar sentences in different languages will be closely mapped in the vector space. The model supports over 50 languages, and does not require specification of the input language.

For each document, $d^r$ contains three dense representations: $d_{1}^r$ corresponds to its body+title, $d_{2}^r$ to its f.p. (first paragraph), and $d_{3}^r$ to its f.p.+title. For the first paragraph, the representation retrieved corresponds directly to the output of the model's encoder, while in the case of the first paragraph + title, mean pooling is performed between the two output vectors corresponding to each component. Finally, for the body + title, the body is segmented into paragraphs and each paragraph is processed individually by the encoder, with the final output vector being obtained through mean pooling of each paragraph's representation and the title. The title and the first paragraph of the documents are used as features with the intuition that sentences in the beginning of a news document usually have the greatest importance to the article \cite{word_importance, automatic_summarization}. Additionally, a representation for the title alone as a feature was not used for the documents due to certain news articles containing only the text of the article and no title.

Regarding the temporal representation, we follow previous approaches \cite{miranda} and expose the temporal representation of a document as the value of its timestamp at the level of the day. When comparing a document's timestamp $d^{ts}$ against a given cluster's timestamp $c^{ts}$, we compute the Gaussian similarity between the two timestamps (with $\mu$ and $\sigma$ corresponding to hyper-parameters) as represented in the function below:

\begin{equation}
score^{ts}(d^{ts}, c^{ts}) = exp \left(-\frac{(d^{ts} - c^{ts}) - \mu}{2\sigma^2}\right)
\end{equation}

Clusters are also divided between dense ($c^r$) and temporal ($c^{ts}$) representations, with each cluster keeping three centroids for each document representation (body+title $c^r_1$, f.p. $c^r_2
$, f.p.+title $c^r_3$) that correspond to the average of the respective representations of each accepted document. When the document's representations are received, each centroid is updated to take them into account. A cluster also maintains timestamps for the newest document ($c^{ts}_1$), the oldest ($c^{ts}_2$), and the mean timestamp between all documents in the cluster ($c^{ts}_3$). To compare two timestamps, we compute the Gaussian similarity between them as proposed in previous work \cite{miranda}, which we refer to as $score^{ts}$. After a cluster is created, it is stored in the cluster pool, a structure that is responsible for maintaining the clusters and archiving old clusters as the system grows in size.

\subsection{Cluster Ranking and Acceptance Models}
\label{rank_merge}

After computing its representations, a given document $d$ is compared against each cluster $c$ in the cluster pool in order to retrieve the most similar cluster to $d$. To determine the similarity between $d$ and each cluster $c$ in the cluster pool, we compute $c$'s ranking score and the best-ranked cluster is then evaluated by the acceptance model. If the cluster is accepted by the model, then $d$ enters the cluster and its representations are updated; otherwise, a new cluster containing $d$ is created. Temporal features are computed through the aforementioned $score^{ts}$ function, and the dense features are obtained through the computation of the cosine similarity ($score^{cos}$), with $d^{r}$ being a given representation of the document and $c^{r}$ a representation of the cluster, as follows:

\begin{equation}
 score^{cos} (d^{r}, c^{r}) = \frac{d^{r} \cdot c^{r}}{|d^{r}| |c^{r}|} 
\end{equation}

The ranking score for a cluster $c$ given a document $d$ and the ranking model's learned SVM weights $u^r$ and $u^{ts}$ is represented as follows:
\begin{equation}
\begin{split}
score^{rank}(d,c) {} & = \sum_{i = 1}^{3}\left( score^{cos}(d_{i}^{r},c_{i}^{r}) \cdot u_{i}^{r} \right) + \sum_{j = 1}^{2}\left( score^{cos}(d_{j+1}^{r},c_{1}^{r}) \cdot u_{j+3}^{r} \right)  \\ & + \sum_{k = 1}^{3}\left( score^{ts}(d^{ts},c_{k}^{ts}) \cdot u_{k}^{ts} \right)
\end{split}
\end{equation}
The ranking model takes the form of a Rank-SVM model \cite{ranksvm}, which we train using a similar scheme to Miranda et al. \cite{miranda}. Given the training partition, each document generates a positive example corresponding to its gold cluster, and 20 negative examples for the 20 best-ranked clusters that are not the gold cluster. These examples are then used in the training of a Rank-SVM to obtain a set of learned weights $w^r$ and $w^{ts}$ for each of the features. After computing the best-ranked cluster $c$ for a given document $d$, the acceptance model determines if the document enters the cluster by computing its acceptance score through an SVM given a bias parameter $b$ and a set of similarity features with learned weights represented by $v^r$ and $v^{ts}$, which takes the following form:
\begin{equation}
\begin{split}
score^{accept}(d,c) {} & = \sum_{i = 1}^{3}\left( score^{cos}(d_{i}^{r},c_{i}^{r}) \cdot v_{i}^{r} \right) + \sum_{j = 1}^{2}\left( score^{cos}(d_{j+1}^{r},c_{1}^{r}) \cdot v_{j+3}^{r} \right) \\ & + \sum_{k = 1}^{3}\left( score^{ts}(d^{ts},c_{k}^{ts}) \cdot v_{k}^{ts} \right) + b
\end{split}
\end{equation}

If $score^{accept}$ is greater than zero, then $d$ is accepted into $c$, otherwise, a new cluster is created and initialized with $d$. The acceptance model is an SVM trained on the training partition of the dataset: each document generates a positive sample for its corresponding gold cluster, and its second-best ranked cluster is given as a negative example.

\subsection{Merging Clusters}

After a cluster receives a new document, we rank its similarity to each of the other clusters in the cluster pool using the ranking model (described in Section \ref{rank_merge}). Each candidate cluster is then evaluated by a third SVM model, which we call \textit{cluster merge model}, and the documents from each cluster with a positive merge decision are inserted into the source cluster.

The intuition for this model is to find separate clusters that have grown to pertain to the same story, and subsequently merge them. This may happen throughout the clustering process, as few documents pertaining to a given story have entered the system, and the acceptance model may mistakenly assign separate clusters to those documents initially. As more relevant documents enter the system, those clusters may end up in similar points in the vector space, and thus should be merged. For this model, we use the eight features specified in Section \ref{rank_merge} as well as $score^{accept}$, and two additional features corresponding to the size of each cluster of the evaluated pair, with the intuition of associating the cluster merging to clusters of small sizes. The $score^{size}$, given a cluster $c$ with $k$ documents and a size limit vector $v$ of length $n$, is represented by the following equation:

\begin{equation}
score^{size}(c, v) = \sum_{i=1}^{n}\left(\begin{cases} \frac{1}{n},  \textnormal{ if }  k > v[i]; \\ 0,  \textnormal{ if }  k \leq v[i] \end{cases} \right)
\end{equation}

To train the model, we generate a dataset by sampling pairs of clusters and labeling them according to whether they should be separated or merged. For each pair, we evaluate the relative F1 given the gold label of each document in the cluster: if the computed value is higher when the clusters are merged, a positive sample is produced and the clusters are merged for the training, otherwise, a negative sample is generated. This is done without forcing documents into their gold clusters, and with the ranking and acceptance model trained accordingly.

\section{Experiments}

\subsection{Dataset}

\begin{table}
\caption{\label{tab:dataset} Statistics for the train and test partitions of the dataset. The training dataset does not contain documents in Slovenian, Croatian, French, Russian or Italian.}
\centering
\small
\setlength{\tabcolsep}{3pt} % Default value: 6pt
\begin{tabular}{c|c|c|c|c|c|c|c|c|c|c}
\hline
& \textbf{Language} & en & es & de & zh & sl & hr & fr & ru & it \\ \hline
\textbf{Train} & \textbf{Docs} & 12233 & 4527 & 4043 & 10 & - & - & - & - & - \\ 
& \textbf{Clusters} & 593 & 416 & 377 & 1 & - & - & - & - & - \\ \hline
\textbf{Test} & \textbf{Docs} & 8726 & 2177 & 2101 & 440 & 37 & 13 & 61 & 231 & 88 \\ 
& \textbf{Clusters} & 222 & 149 & 118 & 9 & 3 & 2 & 2 & 1 & 2 \\ \hline
\end{tabular}
\end{table}

We follow previous work on this task and evaluate our system on a news clustering dataset \cite{rupnik}. Besides the three main languages (English, Spanish and German), this dataset also provides a significant amount of documents in Chinese and Russian, as well as documents in Slovenian, Croatian, French and Italian. These samples allow us to roughly preview the performance of the system on other languages besides the ones it was trained in. The dataset is composed by 34,687 news documents, and it is divided into two sets: a training set comprised of 20,813 articles, and a test set that contains 13,874 articles. The articles in the training set are dated from \textit{18-12-2013} to \textit{02-11-2014}, while the articles in the test set are dated between \textit{02-11-2014} and \textit{27-08-2015}, thus guaranteeing that the articles in the test set are newer and their thematic have not been observed in the training set. Further statistics regarding the dataset are presented in Table \ref{tab:dataset}.

\subsection{Evaluation}

\begin{table*}[]
\caption{\label{tab:monolingual} Results for monolingual clustering on the test dataset.}
\centering
\small
\setlength{\tabcolsep}{3pt} % Default value: 6pt
\begin{tabular}{c|c|ccc|ccc|c}
\hline
\multicolumn{1}{c|}{\textbf{Language}}  & \multicolumn{1}{c|}{\textbf{Systems}} & \multicolumn{3}{c|}{\textbf{BCubed}} & \multicolumn{3}{c|}{\textbf{Standard}} & \multicolumn{1}{c}{\textbf{Clusters}} \\
        & \multicolumn{1}{l|}{}        & \textbf{F1}      & \textbf{P}       & \textbf{R}       & \textbf{F1}        & \textbf{P}       & \textbf{R}       & \multicolumn{1}{l}{}         \\ \hline
        & Miranda et al.               & 92.36   & 94.27   & 90.25   & 94.03     & 98.14   & 90.25   & 326                          \\
 & Staykovski et al.            & 94.41   & 95.16   & 93.66   & 98.11     & 97.60   & 98.63   & 484                          \\
English        & Linger et al.                & 93.86   & 94.19   & 93.55   & \textbf{98.31}     & 98.21   & 98.42   & 298                          \\
        & Saravanakumar et al.         & \textbf{94.76}   & 94.28   & 95.25   & -         & -       & -       & -                          \\
        & Ours                      & 92.43   & 92.76   & 92.10   & 96.46     & 96.50   & 96.41   & 470                          \\ \hline
        & Miranda et al.               & 91.61   & 96.44   & 87.25   & 96.83     & 97.01   & 96.65   & 281                          \\
Spanish & Linger et al.                & \textbf{91.79}   & 93.76   & 90.08   & \textbf{97.68}     & 98.02   & 97.34   & 267                          \\
        & Ours              & 90.39   & 95.01   & 86.20   & 95.48     & 95.48   & 95.48   & 293                          \\ \hline
        & Miranda et al.               & 93.64   & 98.92   & 88.90   & 97.19     & 99.86   & 94.67   & 229                          \\
German  & Linger et al.                & \textbf{94.62}   & 95.13   & 94.31   & 98.70     & 99.16   & 98.24   & 205                          \\
        & Ours               & 93.71   & 97.68   & 90.04   & \textbf{99.07}     & 99.64   & 98.50   & 217                          \\ \hline
\end{tabular}
\end{table*}

Regarding evaluation metrics, we follow the same approach as \cite{dense_vs_sparse,batch_clustering} and report the F1 score and the BCubed F1 \cite{bcubed} score, as well as the associated Precision and Recall scores. Each sample document of the test dataset contains a label with the expected cluster ID, and since the clusters described in the test dataset are monolingual, the crosslingual connections are given by a positive/negative label between two clusters. As such, for the standard F1 score, a \textit{true positive} is described as a pair of two documents with matching cluster labels (monolingual), or a pair of documents whose cluster labels share a positive connection (crosslingual) that have been accepted into the same cluster. A \textit{false positive} corresponds to a pair of documents whose cluster labels do not match (or share a positive connection) that have been accepted into the same cluster. A \textit{true negative} is represented by a pair of documents whose cluster labels do not match (or share a positive connection) that have been accepted into different clusters, and a \textit{false negative} corresponds to a pair of two documents with matching cluster labels (monolingual), or a pair of two documents whose cluster labels share a positive connection (crosslingual) that have been accepted into different clusters.

For the BCubed F1 score, as previously described by Staykovski et al. \cite{dense_vs_sparse} and Amigó et al. \cite{bcubed}, the BCubed precision of a document corresponds to the proportion of documents in its cluster whose cluster label is the same, including itself. The BCubed recall of a document is the proportion of documents with the same label as that document (in the whole dataset) that appear in its cluster. The correctness between two documents $i$ and $j$, given the label $L_i$ and the cluster $C_i$ for each document $i$, is computed as follows:
\begin{equation}
    Correctness(i, j) = \begin{cases} 1, \textnormal{ if } L_i = L_j \textnormal{ and } C_i = C_j  \\ 0,\textnormal{ Otherwise} \end{cases}
\end{equation}

The overall BCubed precision, recall and F1 score are computed as follows:
\begin{equation}
\begin{split}
    \textnormal{BCubed } P &= Avg_i[Avg_{i.C_i=C_j}[Correctness(i,j)]] \\
    \textnormal{BCubed } R &= Avg_i[Avg_{i.L_i=L_j}[Correctness(i,j)]] \\
    \textnormal{BCubed } F_1 &= 2 * \frac{\textnormal{BCubed } P * \textnormal{BCubed } R}{\textnormal{BCubed } P + \textnormal{BCubed } R}
\end{split}
\end{equation}

For the monolingual evaluation, we evaluate the clustering performance of our model on the three main languages of the dataset by performing clustering using only the documents of the specified language, while the crosslingual evaluation uses the entirety of the test set regardless of language. In order to evaluate the results, a gold set of cluster labels is provided for each document that indicates the expected cluster of that document. The clusters are typically multilingual, and in accordance to previous work \cite{miranda, batch_clustering}, the crosslingual evaluation takes into account both monolingual and crosslingual connections between documents of a cluster.

\subsection{Experimental Results}

\begin{table}[]
\caption{\label{tab:crosslingual} Crosslingual clustering results on the test dataset.}
\centering
\small
\setlength{\tabcolsep}{3pt} % Default value: 6pt
\begin{tabular}{l|ccc|ccc|c}
\hline
\multicolumn{1}{c|}{\textbf{Systems}} & \multicolumn{3}{c|}{\textbf{BCubed}} & \multicolumn{3}{c|}{\textbf{Standard}} & \multicolumn{1}{l}{\textbf{Clusters}} \\
\multicolumn{1}{c|}{}        & \textbf{F1}      & \textbf{P}       & \textbf{R}       & \textbf{F1}        & \textbf{P}       & \textbf{R}       & \multicolumn{1}{l}{}         \\ \hline
Miranda et al.               & -   & -   & -   & 84.0     & 83.0   & 85.0   & -                          \\
Linger et al.                & 82.06   & 80.25   & 83.97   & 86.49     & 85.11   & 87.92   & 606                          \\ \hline
(Ours) 4-F Rank+Acc.               & 88.02   & 91.31   & 84.95   & 92.34     & 97.26   & 87.09   & 957                          \\
(Ours) 8-F Rank+Acc.                & 89.24   & 92.62   & 86.11   & 93.76     & 97.66   & 90.15   & 1023                          \\
(Ours) 8-F Rank+Acc.+M.                         & \textbf{90.10}   & 89.70   & 90.51   & \textbf{97.21}     & 97.01   & 97.42   & 812                          \\ \hline
\end{tabular}
\end{table}

As shown in Table \ref{tab:monolingual}, for the monolingual evaluation our system is on-par with Miranda et al.'s in English and German on both metrics, but is surpassed by Linger et al.'s in all languages and modularities except for German when evaluating both metrics.

For crosslingual clustering, as shown in Table \ref{tab:crosslingual}, our system achieves state-of-the-art performance on BCubed F1 \cite{bcubed} (+8.04) and on the standard F1 (+11.33) despite producing a larger amount of clusters. Furthermore, we perform an ablation study that shows the relative importance of the system components. 4-F Rank+Acc. refers to the clustering system with a 4-feature ranking and acceptance model, which used only $score^{cos}(d^{r}_1, c^{r}_{1})$ and the timestamp features. Adding the other features (8-F Rank+Acc.) improved both standard (+1.42) and BCubed F1 (+1.22). Finally, the cluster merge model was added to our system (8-F Rank+Acc.+M.), resulting in gains for both standard (+3.35) and BCubed F1 (+0.86).

\begin{table}[]
\caption{\label{tab:other_langs} Clustering results on other languages.}
\centering
\small
\setlength{\tabcolsep}{3pt} % Default value: 6pt
\begin{tabular}{l|ccc|ccc|c}
\hline
\multicolumn{1}{c|}{\textbf{Languages}} & \multicolumn{3}{c|}{\textbf{BCubed}} & \multicolumn{3}{c|}{\textbf{Standard}} & \multicolumn{1}{l}{\textbf{Clusters}} \\
\multicolumn{1}{c|}{}        & \textbf{F1}      & \textbf{P}       & \textbf{R}       & \textbf{F1}        & \textbf{P}       & \textbf{R}       & \multicolumn{1}{l}{}      \\ \hline
Chinese             & 96.18  & 100.00   & 92.65   & 99.07     & 100.00   & 98.16   & 28                     \\
Slovenian             & 76.92   & 100.00   & 62.50   & 79.67     & 100.00   & 66.21   & 12                   \\
Croatian             & 77.85   & 100.00   & 63.73   & 74.99     & 100.00   & 60.00   & 5                   \\
French             & 98.50   & 100.00   & 97.04   & 99.69     & 100.00   & 99.39   & 3                 \\
Russian             & 100.00   & 100.00   & 100.00   & 100.00     & 100.00   & 100.00   & 1               \\
Italian             & 98.86   & 100.00   & 97.75   & 98.78     & 100.00  & 97.59   & 3               \\ \hline
\end{tabular}
\end{table}

Given the nature of our system, we evaluated it on the remaining languages of the dataset as shown in Table \ref{tab:other_langs}. Our ranking, acceptance and cluster merge models were not trained on any data from these languages (with the exception of Chinese), making this a zero-shot clustering scenario. Chinese, French, Russian and Italian document clustering had high F1-scores, with results above 95\%, and both Slovenian and Croatian had initial clustering scores above 70\%.

\section{Conclusion}

We presented a clustering model that produces state-of-the-art results at a multilingual level without depending on language-specific features, and that maintains quality at a monolingual level on-par to previous work on news clustering. 
\begin{comment}
Our model was evaluated on English, German and Spanish, as well as on a significant amount of documents in Chinese, Croatian, Slovenian, French, Russian and Italian despite not seeing these languages during training.
\end{comment} 
We demonstrated that it is possible to improve results by utilizing contextual embeddings to represent documents at a crosslingual level, and how a linear SVM can be trained in order to perform such a task. 
By reducing the complexity of the clustering space, we motivate future research on topics such as clustering while taking user feedback into account, and high-performance vector search to improve clustering speed and scalability. Our system also enables computational efficiency improvements by allowing most operations to be paralellized. We make our code available as open-source\footnote{\textit{https://github.com/Priberam/projected-news-clustering}}.

\section*{Acknowledgements}

This work is supported by the EU H2020 SELMA project (grant agreement No 957017).

\end{document}